# KPI-Conditioned Generative Design of Automotive Hood Inner Panels: A Two-Stage Retrieval–Generation Pipeline with Surrogate-Based Performance Estimation

Sudeep Chavare

*Independent researcher*

## Abstract

An inner hood panel must meet a deflection target, stay below a stress limit, and hit a mass target. Machine-learned surrogates have made the forward direction, geometry to performance, fast and routine. The inverse direction, producing geometry from a stated requirement, remains largely unaddressed for industrial parts whose design space is organized into discrete topology families rather than a continuous parameterization.

This work presents a two-stage pipeline for that inverse problem. A reachability stage determines which topology families can satisfy a given requirement vector. A conditional variational autoencoder then generates point-cloud geometry within a selected family, and a neural-operator surrogate estimates the performance of each candidate. The pipeline is built entirely from public data and freely available compute, and is deployed as an interactive tool.

The pipeline works, with qualifications that are reported as primary findings rather than caveats. The surrogate is accurate in aggregate, but its error is comparable to the performance differences it is asked to discriminate, which bounds what can be claimed for any individual generated design. That ratio of surrogate error to within-class signal is argued to be the quantity that determines whether a pipeline of this kind can work at all.

**Keywords:** generative design; inverse design; surrogate modeling; neural operators; point clouds; automotive structures; uncertainty of surrogate resolution

## 1. Introduction

An inner hood panel must satisfy a deflection target under hood-lift load, remain below a stress limit, and meet a mass target. The engineer receives these as requirements; what must be produced is a rib-and-cutout architecture. The conventional workflow proceeds by selecting a carryover panel or an existing concept, building CAD, meshing, solving, evaluating, and iterating. The expensive step in that loop is not the solve, but the selection and first-geometry construction that precede it.

Machine learning surrogates have made the forward direction fast. Transformer-based neural operators such as Transolver [1] and its geometry-aware extension GeoTransolver [2] predict full transient deformation fields for industrial crash models in milliseconds rather than hours [3, 4]. The inverse direction, specifying performance and obtaining geometry, remains substantially harder, and the difficulty is not only computational.

### 1.1 Inverse design is underdetermined at the topology level

Let the requirement vector be $k = (\sigma_max, \delta_max, m)$ and let G denote the space of manufacturable hood geometries. The forward map $f: G \rightarrow R^3$ is well defined: one geometry yields one performance triple. Its inverse is not. Empirically, a feasible requirement triple is satisfied by a mean of 3.4 distinct topology families across this dataset (Section 4.2).

More consequentially, the information required to choose among those families is absent from k. Stamping draw depth, hinge and latch packaging, pedestrian headform impact zones, platform carryover, tooling reuse, and assembly interfaces all bear on topology selection, and none appear in a structural KPI vector or in any available dataset. Two topologies can be KPI-identical and manufacturing-incompatible. A model conditioned only on k therefore cannot correctly select the family. This is not because it is undertrained; the deciding variables are simply not among its inputs.

This is the architectural justification for the two-stage split adopted here: the model narrows the topology space to structurally reachable candidates, and the engineer applies the constraints the model cannot observe.

### 1.2 Contributions

- A variability decomposition of an industry-grade hood design space establishing that topology selection dominates parametric variation by a factor of 3 to 7, and that this dominance is graded across KPIs: mass is most locked by topology, stress least.
- A reachability formulation for topology-family retrieval, with per-family and per-KPI resolvability labeling derived from comparing surrogate error against within-family spread.
- A conditional generative model for hood geometry that outperforms KPI-based retrieval and attains 68% of the retrieval-oracle bound, improving on the reference design in 103 of 104 families.
- Three quantified negative results: surrogate-in-the-loop conditioning collapse with its mechanism; the reconstruction resolution limit; and tangential dominance in Chamfer-trained point-cloud generation at 6.2:1, which explains the other two.
- A deployed tool that reports performance estimates only at the resolution its evidence supports.

## 2. Related Work

The CarHoods10k dataset [6] was released explicitly to support representation learning and design optimization in engineering applications, and its authors demonstrated three use cases: geometric deep learning for a compact latent representation, machine learning models predicting mechanical performance from that representation, and integration into an evolutionary topology optimization approach under manufacturability constraints. The second use case extends to generation: latent codes were optimized by differential evolution against the metamodel and decoded through the autoencoder, reducing predicted maximum stress by 45.9% on average, with the decoded point clouds assessed by visual inspection [6]. Latent representation learning, forward metamodeling, performance-driven search, and generation guided toward a performance extremum are therefore all established on this dataset. Two things remain unestablished. The first is generation conditioned on a specified requirement vector, meeting a stated stress, deflection, and mass rather than minimizing them. The second is any assessment of whether a generated design attains the performance attributed to it.

AutoHood3D [7] builds directly on the same hood lineage, taking 100 inner panels from CarHoods10k [6] as base shells and adding a convex-hull outer envelope to form a dual-shell geometry, then generating 16,000-plus variants for a rotary-dip-painting fluid–structure interaction load case. Its geometry generation is procedural rather than learned: cutout curves extracted from the base panels are clustered by perimeter and area, then sampled from those clusters under symmetry and spacing constraints. Its machine learning contribution is forward regression, benchmarking MLP, PointNet, GraphSAGE, Graph U-Net and PointGNNConv architectures against normalized deformation. A paired text–point-cloud corpus for language-conditioned shape synthesis is released, but conditioned on geometric descriptors (spacing, offset, curve count) rather than on performance, and without a trained model. The load case, the conditioning signal, and the generative mechanism all differ from the present work.

Point-cloud generative frameworks for inverse design have been demonstrated in metamaterials, where a latent space organized by mechanical properties clusters naturally by unit-cell type and supports property-guided generation. The architectural parallel to the present two-stage formulation is close, with unit-cell type playing the role of topology family; the distinction here lies in the discrete, industry-validated family structure and in the explicit treatment of surrogate resolution limits.

## 3. Data and Geometry Representation

### 3.1 Source dataset

CarHoods10k [5, 6] provides car hood inner frames generated by the automated CATIA v5 workflow of Ramnath et al. [13]. Base geometries were formed by pairing each of ten idealized hood skins with each of eleven pocket feature patterns, and each pairing was varied over one hundred parameter settings, giving 11,000 nominal designs of which 10,478 were generated successfully [6]. The released files label these pairings skin_N and the deposit contains 109 of them; each is treated here as a topology family. Note that skin in [6] denotes the hood surface, of which there are ten, rather than the surface-and-pattern pairing, so the term family is used throughout in preference to the file-naming convention. Performance values come from Ansys finite element analysis under hood-lift and hood-twist load cases representing driving conditions; the two correlate strongly, so only hood-lift results are released [6]. The three reported quantities are maximum equivalent (von Mises) stress, maximum directional z-axis deformation, and geometry mass, obtained under a boundary condition setup held uniform across geometries, with 10,070 of the 10,478 geometries returning successful analyses [6]. Both the geometries and their performance values were checked by industry experts and benchmarked against real-world hood designs, establishing validity, realism, manufacturability, and sufficient design variability [6]. The released meshes are watertight with properly oriented surface normals, and are of genus one or higher [6]. The dataset is released under a CC0 1.0 public domain dedication, imposing no restriction on use, modification, or redistribution.

### 3.2 Integrity audit and pairing verification

Two independent failure modes exist in the raw data: CAD failures, in which the STL is absent, and finite element failures, in which the results row is absent. These do not coincide, so the usable set is their intersection. Critically, the OutputNumber column is a geometry identifier rather than a row index: failed rows are omitted rather than blanked, so any positional join shifts every subsequent pairing. Joining explicitly on OutputNumber and auditing yielded 9,867 matched pairs from 10,070 results rows and 10,433 geometry files, with no duplicate identifiers and no rows failing physical sanity bounds.

Subsequent filtering removed 51 zero-byte STL files (38-byte headers containing no triangles, cleanly separated from the next-smallest file at 20 MB) and families with fewer than 40 usable variants. One family had no geometry files; another retained 24 of 96. The final training corpus comprises 9,479 designs across 104 families, with a median of 95 variants per family.

Automated checks cannot detect a consistent off-by-one error. Pairing was therefore verified physically. The meshes are watertight solids, so mass follows from geometry as $m = \rho V$ with $\rho = 7850$ kg/m³. For the first design tested, the geometric mass was 14.86 kg against a manifest value of 14.86 kg, agreement to five significant figures that confirms both the pairing and the material. Across 20 designs from different families the mean absolute error was 1.5% with a positive bias, consistent with the reported mass deriving from a shell model of nominal thickness while the geometric computation measures the CAD solid, since mid-surface extraction loses material at flanges, hems, and fillets. An important consequence follows: with approximately 3% scatter, geometric mass cannot resolve within-family differences of 1.89%, and functions as a gross-failure detector rather than a measurement.

### 3.3 Point-cloud representation

STL triangle counts vary per design and carry no consistent ordering, so no fixed tensor shape exists. Triangle connectivity is a meshing artifact rather than design intent: two identical hoods remeshed differently produce different triangles and the same shape. Point clouds are permutation-invariant and fixed-size, and Chamfer distance compares two point sets without requiring correspondence, which matters because a decoder's output ordering bears no relation to its input's.

Peak stress is a local phenomenon, concentrating at fillets, rib terminations, and cutout corners; uniform surface sampling under-resolves precisely these regions. Hybrid sampling was therefore used: 30% uniform surface samples capturing the macro envelope, and 70% curvature-weighted samples drawn from a four-fold oversampled pool with probability proportional to the mean curvature magnitude raised to the power 1.5, measured at a radius of 4 mm, tight enough to detect rib and cutout edges rather than broad panel curvature. Farthest point sampling then reduced the pool to a fixed 16,384 points, deduplicating and guaranteeing spatial coverage.

### 3.4 Normalization

Per-hood normalization to a unit bounding box destroys scale. Because mass is a direct function of material volume, stripping scale would render mass unlearnable. A single global affine transform was therefore applied across the entire dataset:

$$\hat{x} = \frac{x-c}{s}, \quad c = (-335.7,\ 0.0,\ 672.2)\ \text{mm}, \quad s = 886.1\ \text{mm} \tag{1}$$

Verification confirmed the choice: the per-hood normalized diagonal retained a coefficient of variation of 0.051, non-zero and therefore preserving relative size. The exactly zero y-component of the center also confirms that the dataset's hoods are consistently positioned and symmetric about the vehicle centerline.

### 3.5 Sampling resolution

Representation density was selected by measuring signal-to-noise directly, as the Chamfer distance between the two most extreme designs within a family divided by the distance between two independent subsamples of the same design.

| Points per design | Self-distance (noise) | Extreme-pair distance (signal) | SNR |
|---|---|---|---|
| 4,096 | 0.0302 | 0.0381 | ≈ 1.2 : 1 |
| 8,192 | 0.000070 | 0.019 | 272 : 1 |
| 16,384 | 0.000073 | 0.0166 | 228 : 1 |

*Table 1. Signal-to-noise of the Chamfer metric as a function of sampling density.*

The 4,096-point configuration was discarded. At that density the mean nearest-neighbor spacing of approximately 20 mm exceeds the geometric differences being measured, so the Chamfer objective measures sampling noise rather than shape. This diagnosis accounts for an early training failure in which the loss remained flat and the latent variable collapsed (Appendix A).

### 3.6 Data splits

Splits are stratified within families: 70% training, 10% validation, and 20% test for each family, so every family appears in every split. This tests generalization to unseen variants of a known family, which is what the deployed tool performs. It does not test generalization to unseen topologies, and that is stated as a scope limit rather than presented as a gap. The identical split is shared by the surrogate and the generative model, so that surrogate-based evaluation of generated designs never involves a design the surrogate was trained on. Verified KPI means and standard deviations agree across the three splits to within 1.5 MPa in stress and 0.13 mm in deflection.

## 4. Design Space Characterization

### 4.1 Variability decomposition

Before selecting an architecture, the variability structure of the design space was characterized. For each KPI the coefficient of variation was computed within each family and averaged across families, and separately across the family means:

$$\mathrm{CoV}_{\mathrm{within}}(k) = \frac{1}{F}\sum_i \frac{\mathrm{sd}(k|\text{family } i)}{\bar{k}_i} \tag{2}$$

$$\mathrm{CoV}_{\mathrm{across}}(k) = \frac{\mathrm{sd}(\{\bar{k}_1,\ldots,\bar{k}_F\})}{\bar{k}} \tag{3}$$

| KPI | Within-family CoV | Across-family CoV | Ratio |
|---|---|---|---|
| Maximum stress | 8.75% | 27.94% | 3.19 |
| Maximum deflection | 5.80% | 41.31% | 7.13 |
| Mass | 1.89% | 14.12% | 7.48 |

*Table 2. Variability decomposition over 108 families and 10,070 finite element results.*

All three ratios substantially exceed unity, but the ordering carries the engineering content. Mass is nearly locked by topology at 1.9% within-family variation: once the family is chosen, mass is determined to within a few percent. Deflection is set by topology with modest trim. Stress retains the greatest within-family freedom, and its across-to-within ratio is less than half that of the other two KPIs. The resulting claim, which shaped the remainder of this work, is that topology selection determines mass, sets the stiffness regime, and bounds stress, while feature-level variation then tunes stress with secondary influence on deflection and almost none on mass. This decomposition is consistent with the dataset authors' own observation that pocket feature patterns dominated hood performance relative to other features, and that varying those patterns was what introduced sufficient performance variation into the corpus [6]. Since a family here is exactly a skin-and-pocket-pattern pairing, the dominance of the family term is the same effect measured on the assembled dataset rather than during its construction.

Individual families make the effect concrete. In one family, deflection spans 5.21 to 5.38 mm (3.2% of its mean) while stress spans 173.7 to 248.4 MPa, or 38.8% of its mean, across the same 98 designs. Global stiffness is a property of the rib architecture; peak stress is a property of where that stiffness concentrates, at fillet radii, cutout proximities, and rib terminations. Two implications follow: the substantive task of the generative stage is stress control, and the two-stage split mirrors the underlying mechanics rather than being merely a modeling convenience. Figure 1 shows the discrete character of the topology variation.

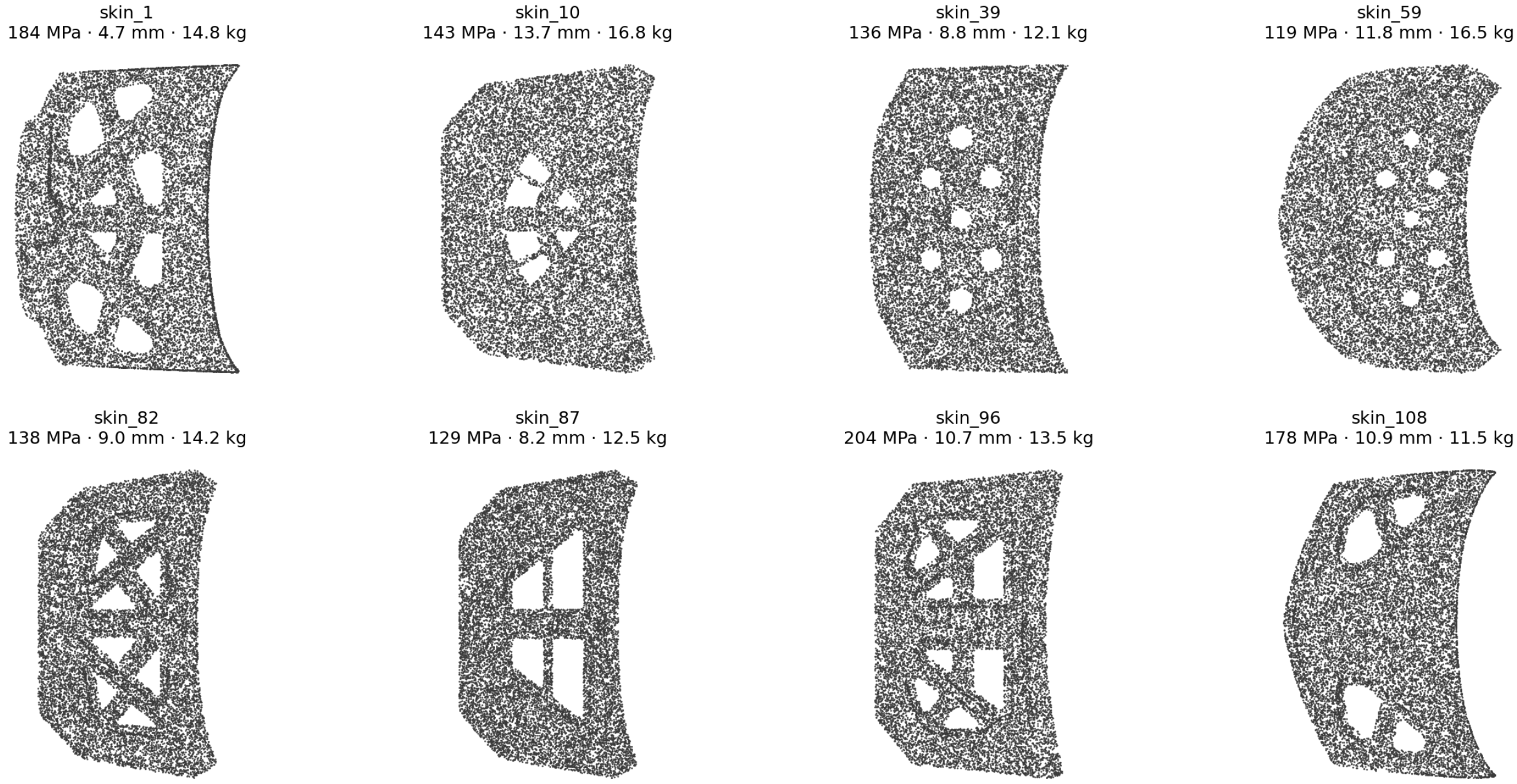

*Figure 1. Eight of the 104 topology families, each shown as a point cloud of one member and labeled with its maximum stress, maximum deflection, and mass. Rib and cutout architecture changes discretely between families rather than varying continuously, and the labeled KPI triples span much of the corpus range. This is the across-family variation quantified in Table 2.*

## 4.2 Correlation structure

Family-mean correlations are 0.506 between stress and deflection, 0.124 between deflection and mass, and −0.121 between stress and mass. Mass is therefore effectively decoupled from both structural KPIs at the family level. Within this design space, added mass is not purchasing stiffness; rib architecture and material distribution are doing that work. Lightweighting opportunities consequently exist across the stiffness range, and the binding constraint on requirement feasibility is the stress–deflection coupling rather than a mass–stiffness trade.

## 5. Methods

### 5.1 Stage one: reachability retrieval

For each family the achievable envelope and mean are precomputed per KPI. Given a target t, the fractional shortfall for family i and KPI k is

$$e_i^k = \frac{\max\left(k_i^{\min} - t^k, t^k - k_i^{\max}, 0\right)}{t^k} \tag{4}$$

which vanishes when the target lies inside the family's envelope. A family qualifies when the shortfall is at most τ = 0.03 for all three KPIs. Tolerance is defined relative to the target value, following engineering convention: within 3% of a 10 mm target means ±0.3 mm. Figure 2 shows the envelopes over which this test is evaluated.

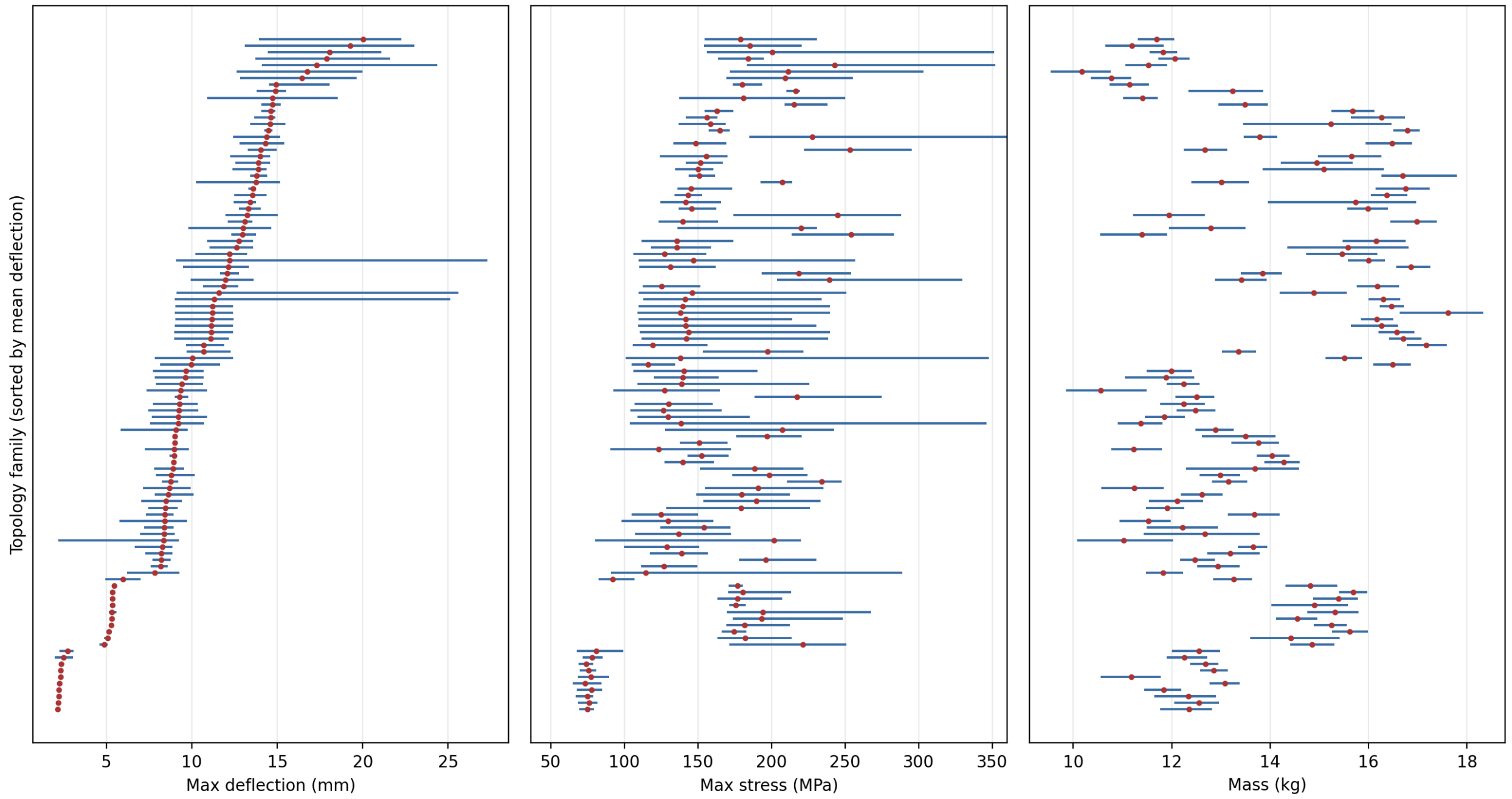

*Figure 2. Achievable KPI envelope of each topology family, ordered by mean deflection. Bars span the observed minimum to maximum within a family and markers give the family mean, so bar length is the within-family spread and the scatter of markers is the across-family spread. Retrieval scores a requirement vector against these envelopes. The ordering is shared across all three panels, so the near-monotone deflection panel set against the scattered stress and mass panels is the weak family-level correlation reported in Section 4.2.*

Strict filtering alone proved insufficient in deployment, because the generative model returns approximately the family mean (Section 6.3). A family whose envelope merely contains the target may have a mean far from it. Ranking therefore combines shortfall with centrality, weighted by a user-selected primary requirement:

$$S_i = \sum_k w^k \left[ e_i^k + \lambda_c \left| \frac{\bar{k}_i - t^k}{t^k} \right| \right], \quad \lambda_c = 1, \quad w^k \in \{1,3\} \tag{5}$$

The centrality term routes around the compression identified in Section 6.3 rather than attempting to correct it.

### 5.2 Surrogate model

The surrogate performs global scalar regression rather than field prediction, mapping $R^{N\times3}$ to $R^3$. A Transolver trunk [1, 2], in the NVIDIA PhysicsNeMo implementation [15], is followed by concatenated mean and max pooling over output tokens and a multilayer perceptron head:

$$h = \text{Transolver}(X), \quad z = \left[\max_i h_i \parallel \frac{1}{N}\sum_i h_i\right], \quad \hat{k} = \text{MLP}(z) \tag{6}$$

The configuration uses four layers, hidden width 64, four attention heads, 32 physics-attention slices, and an output token width of 32, totalling 0.22 M parameters. Targets are standardized on the training split; the objective is a smooth L1 loss optimized with AdamW at a learning rate of $5\times10^{-4}$ under cosine annealing. The per-token path is retained deliberately, so that field prediction becomes a small extension rather than a rewrite.

### 5.3 Stage two: conditional generation

Within a family, all designs share topology, the same rib pattern and cutout layout, differing only in feature dimensions. The model therefore need not invent topology; it must deform a known one. Letting $P_{tpl}$ denote a fixed reference design for the family, taken as the median-deflection member of the training split, the decoder predicts a displacement field:

$$\hat{P} = P_{\text{tpl}} + D\left(P_{\text{tpl}}, z, c\right), \quad c = \left[\tilde{k} \parallel e_{\text{skin}}(i)\right] \tag{7}$$

where $\tilde{k}$ is the standardized KPI target and $e_{skin}(i)$ is a 16-dimensional learned per-family embedding. A single model serves all 104 families. With approximately 21 training designs per family, per-family models would be untrainable; the shared model trains on 6,606 designs, learns the general operation of deepening ribs to reduce deflection, for instance, and specializes through the embedding. The objective combines Chamfer distance with the Kullback–Leibler term of the variational autoencoder [9], in the conditional form of Sohn et al. [10]:

$$L = d_{CD}\left(\hat{P}, P_{\text{tgt}}\right) + \beta\, D_{KL}\left(q(z \mid P, c) \parallel \mathcal{N}(0, I)\right) \tag{8}$$

$$d_{CD}(A, B) = \frac{1}{|A|}\sum_{a\in A} \min_{b\in B} \|a - b\| + \frac{1}{|B|}\sum_{b\in B} \min_{a\in A} \|a - b\| \tag{9}$$

The decoder uses group normalization with GELU activations throughout, and its output layer is initialized with small random weights rather than exact zeros. Both choices were forced by training failures documented in Appendix A.

## 6. Results

### 6.1 Retrieval calibration

| Target sampling | At least one exact family | Mean families | No family |
|---|---|---|---|
| Real designs (n = 300) | 100% | 7.3 | 0% |
| ±10% perturbed | 96% | 5.9 | 4% |
| Uniform over marginal ranges | 12.8% | n/a | 87% |

*Table 3. Retrieval behavior at a 3% tolerance under three target-sampling regimes.*

The uniform-sampling result is not a failure of the method. The achievable set is a thin curved sheet within the marginal bounding box, so most random triples request physically incompatible combinations. The realistic-target rows are the operative ones, and they show τ = 0.03 to be well calibrated: it never starves on feasible requests while still filtering 104 families to approximately seven.

### 6.2 Surrogate accuracy

A PointNet baseline [8] of approximately 1.4 M parameters was trained first for comparison, using a shared per-point multilayer perceptron with symmetric pooling.

| Configuration | Parameters | Stress | Deflection | Mass |
|---|---|---|---|---|
| PointNet, 8k pts, 3,120 designs | 1.4 M | 8.34% | 7.40% | 2.50% |
| Transolver, 8k pts, 3,120 designs | 0.22 M | 9.09% | 7.12% | 2.66% |
| Transolver, 16k pts, 9,479 designs, 80 ep | 0.22 M | 8.14% | 6.46% | 2.36% |
| Transolver, 16k pts, 9,479 designs, 250 ep | 0.22 M | 7.34% | 5.70% | 1.96% |

*Table 4. Test mean absolute percentage error across architecture, resolution, and dataset scale.*

Two architecturally dissimilar models converged to within 0.75 percentage points of one another despite a six-fold parameter difference, and both plateaued on stress by epoch 35 while deflection continued to improve. Architecture was therefore not the binding constraint. Increasing sampling density and dataset size, with training extended to convergence, reduced stress error by 19%, deflection by 20%, and mass by 26% relative to the initial configuration. The apparent nine-percent plateau was a data limit rather than an information-theoretic one. Coefficients of determination at the final configuration are 0.826 for stress, 0.946 for deflection, and 0.963 for mass. Figure 3 gives the corresponding parity plots.

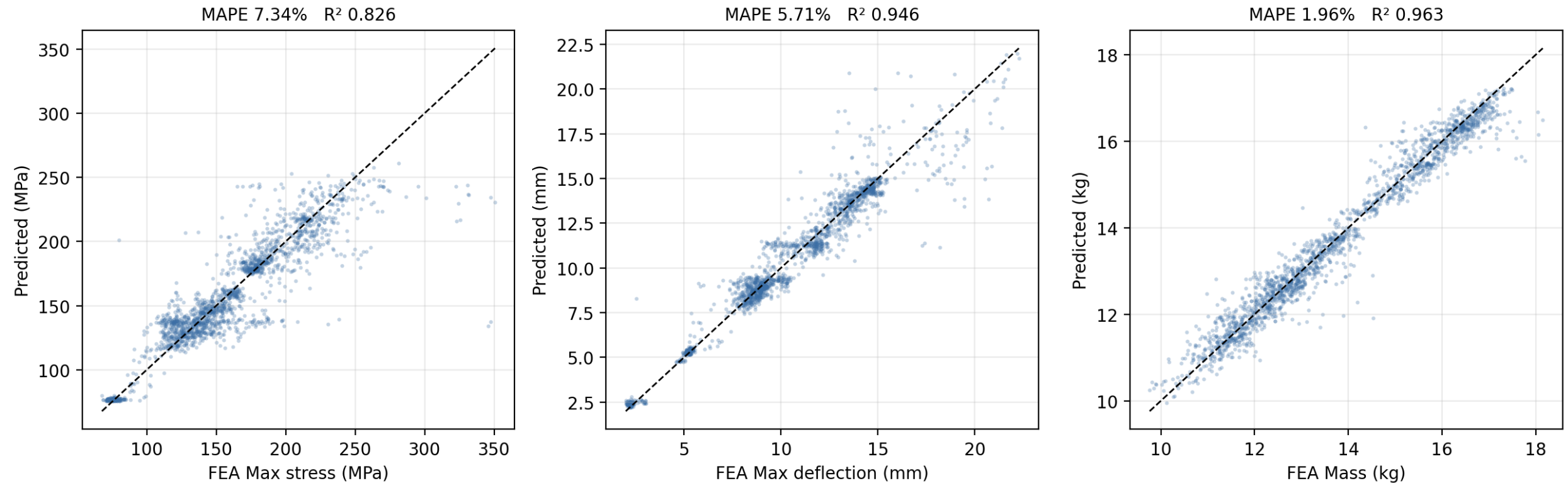

*Figure 3. Test-set parity for the final surrogate configuration (Transolver, 16,384 points, 9,479 designs, 250 epochs). Dashed lines mark exact agreement. Deflection and mass track the diagonal closely across the full range; stress scatter widens above roughly 250 MPa, where designs are sparse, and that region dominates the higher stress error.*

### 6.3 Generative model performance

| Configuration | Chamfer distance | Equivalent (mm) |
|---|---|---|
| Reference design (no generation) | 0.0233 | 20.6 |
| Nearest-KPI retrieval | 0.0199 | 17.7 |
| Conditional autoencoder (this work) | 0.0180 | 16.0 |
| Retrieval oracle (best real family member) | 0.0168 | 14.9 |

*Table 5. Validation Chamfer distance against retrieval baselines and the oracle bound.*

The model outperforms KPI-based retrieval by 1.7 mm and attains 68% of the interval between the reference design and the oracle bound, confirming that it deforms rather than memorizes. Per-family evaluation across all 104 families shows 103 improving on their reference design, a median improvement of 8.0%, and a median KPI sensitivity of 8.60 mm, with no family below 0.5 mm. Sensitivity is measured as the mean per-point geometric difference between generating at a family's minimum and maximum KPI values; at 8.6 mm against approximately 16 mm of total variation, the conditioning demonstrably drives geometry. The single regressing family has four validation samples and lies within noise. Figure 4 contrasts a family reference design with a design generated from it.

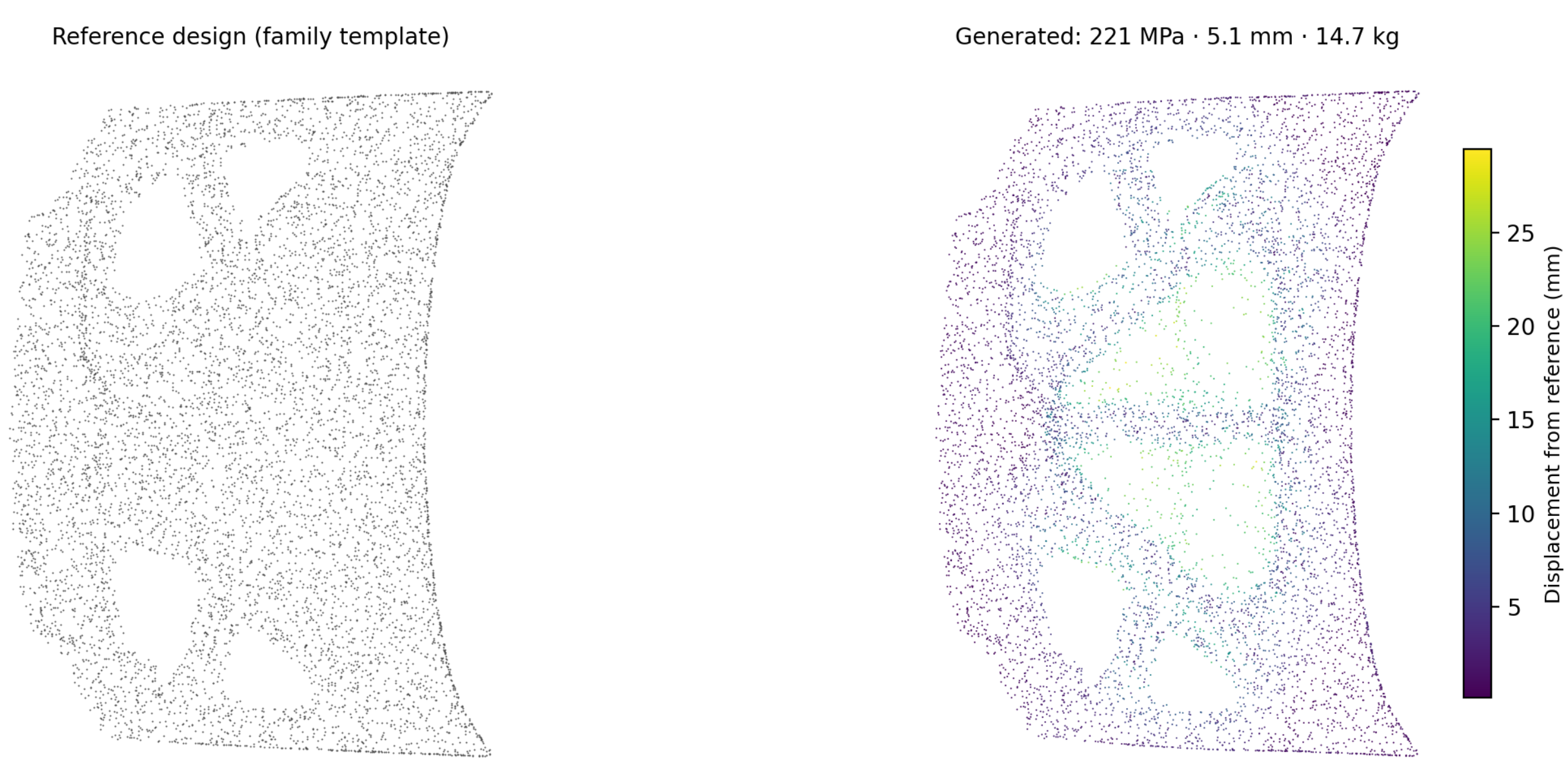

*Figure 4. Family reference design (left) and a design generated from it under a KPI target (right), the latter colored by distance from the reference. Change concentrates in the interior rib and cutout region while the outer envelope is nearly unchanged. Section 7.3 shows this displacement to be predominantly tangential to the surface, at a ratio of 6.2:1 against the normal component.*

*KPI compression*

| Requested stress (MPa) | Achieved stress (MPa) |
|---|---|
| 183.3 | 202.8 |
| 221.4 | 222.0 |
| 245.4 | 223.6 |

*Table 6. Requested versus surrogate-estimated stress across one family's range.*

The model tracks mid-range requests accurately and compresses the extremes toward the family mean, covering approximately 34% of the requested span. This is regression to the conditional mean under thin per-family data, with extreme-KPI members the rarest. The deployed system mitigates rather than solves this, through the centrality term of Section 5.1 and by ranking final candidates on achieved rather than requested performance.

*Latent collapse*

The Kullback–Leibler term fell to zero in every run at every value of $\beta$ tested, including $10^{-5}$. This is expected rather than pathological: once family and KPI are both fixed, the conditional distribution is near-deterministic and the latent variable has nothing to encode. Three latent samples produced stress estimates of 221.4, 221.6, and 221.4 MPa. The deployed model is therefore accurately described as a conditional autoencoder rather than a variational one: the first stage removed the multimodality that the latent was intended to represent.

## 7. Negative Results

### 7.1 Surrogate resolution limits per-design prediction

| KPI | Surrogate MAPE | Mean within-family CoV | Resolvable families |
|---|---|---|---|
| Stress | 7.34% | 8.75% | 47 / 104 (45.2%) |
| Deflection | 5.70% | 5.80% | 44 / 104 (42.3%) |
| Mass | 1.96% | 1.89% | 37 / 104 (35.6%) |

*Table 7. Surrogate error against within-family spread. A KPI is resolvable for a family when its within-family coefficient of variation exceeds the surrogate error.*

High coefficients of determination are achieved on across-family variance, which is three to seven times larger than within-family variance; they do not imply within-family discrimination. Of the 104 families, 29 resolve no KPI, 33 resolve one, 31 resolve two, and 11 resolve all three. Within-family stress variation ranges from 0.99% to 29.72%, a thirtyfold spread, so the statement that the surrogate cannot resolve within-family differences is false as a blanket claim and true for specific families. The deployed tool reports which is which. For the 29 families resolving nothing, the members are near performance-identical, and the honest answer is that any member lands in the same place, which remains actionable.

### 7.2 Surrogate-in-the-loop conditioning collapses

To address KPI compression, the model was retrained at 16,384 points with an auxiliary term penalizing the frozen surrogate's prediction of the generated cloud:

$$L = d_{CD}\left(\hat{P}, P_{\text{tgt}}\right) + \lambda \left\| \frac{S(\hat{P}) - \mu^k}{\sigma^k} - \tilde{k} \right\|^2, \quad \lambda = 0.5 \tag{10}$$

Chamfer distance improved to 0.0160, or 14.2 mm, below the retrieval oracle. KPI conditioning, however, collapsed entirely: the geometric difference between generating at a family's minimum and maximum KPI values fell from 8.60 mm to 0.0106 mm. Generation became independent of the requested KPI, producing identical output to four decimal places across the full requested range.

The mechanism is diagnosable. The surrogate's stress error of 7.34% is of the same order as the 8.75% mean within-family stress spread it was being asked to supervise, and exceeds that spread in 57 of the 104 families (Section 7.1), so the auxiliary gradient was largely noise. Against a noisy target, the loss-minimizing strategy is to predict the conditional mean and ignore the condition, which is exactly what the model learned. Validation KPI loss plateaued at approximately 0.10, the error floor of always predicting the center. Stated generally: surrogate-in-the-loop conditioning collapses when surrogate error exceeds the within-class signal it supervises. This is the same resolution limit as Section 7.1, manifesting as a training failure rather than an evaluation one.

### 7.3 Mesh reconstruction is blocked by the generation objective

Four reconstruction routes were attempted: Poisson surface reconstruction [11], ball pivoting [12], alpha shapes, and template deformation. All were evaluated on real point clouds, the optimistic case, with no contribution from generative error.

| Method | Outcome |
|---|---|
| Poisson (depth 9) | Fills cutouts; enclosed volume 445 kg against 11.8 kg true. Surface area 62–75% of original: an envelope, not a shell. |
| Ball pivoting | Follows the surface but recovers 41% of true area (±0.01 across designs), reconstructing one face of a 1.4 mm shell. Output visibly perforated. |
| Alpha shape | Best surface method; area ratio rises from 0.02 to 0.96 as α increases from 6 to 60 mm, but face count falls above α ≈ 22 mm, indicating gap-spanning. Ribs and cutouts read correctly at 22 mm; surface remains lace-like. |
| Template deformation | Clean inherited topology, but inverse-distance interpolation across sharp creases rounds every edge. Clamping preserves crispness only by reducing displacement to 0.3 mm, i.e. by returning the template. |

*Table 8. Reconstruction routes evaluated on real point clouds.*

The proximate cause is sampling density: mean nearest-neighbor spacing at 16,384 points is 4.7 mm on a panel of approximately 2.6 m², against a shell thickness of 1.4 mm and fillet radii of similar order. No interpolation scheme recovers geometry that was never sampled.

The deeper cause emerged from decomposing the generated displacement against template surface normals. Writing $d = \hat{P} - P_tpl$ and n for the unit surface normal:

$$d_\perp = d \cdot n, \quad d_\parallel = \|d - (d \cdot n)\, n\| \tag{11}$$

| Component | Mean magnitude |
|---|---|
| Normal, $|d\perp|$ | 2.78 mm |
| Tangential, $d\|$ | 17.37 mm |
| Ratio | 6.2 : 1 |
| Signed normal mean | +0.01 mm |

*Table 9. Decomposition of the generated displacement field against template surface normals.*

The signed normal mean of essentially zero is the decisive figure: outward and inward motion cancel, so the net surface motion is nil. The model predominantly redistributes points across the surface rather than deforming it. A mechanism consistent with this is the objective itself: Chamfer distance is minimized by placing output points near the target surface and is indifferent to which point goes where, so tangential sliding reduces the loss as effectively as normal deformation and is easier to learn. This attribution is not established here, however, because the same decomposition was not computed between the template and real members of a family. Within a family the topology is fixed and only feature dimensions vary, so resizing an opening on a fixed skin is itself an in-plane operation; a tangential-to-normal ratio of this order may therefore be a property of the design space rather than an artifact of the loss. Distinguishing the two requires the ratio measured on real within-family pairs, which is left to future work.

Two further observations bound what the generator does produce. The decoder is residual and preserves cardinality, so every output point is a template point that has moved. The model can neither add nor delete material, and cannot introduce a cutout absent from the template. Any apparent topological change in generated output is therefore necessarily a density vacancy left by points migrating away, which is also why reconstruction routes that read a sampling void as a boundary carve holes through continuous surface. Comparing signed depth change against the template for one family, the generated design attains a mean absolute depth change of 3.2 mm, where real members of that family differ from their template by 6.5 mm on average. The model deforms at roughly half the amplitude the data exhibits. The generated design is a conservative but plausible family member rather than an undeformed copy, and this is the same regression toward the family mean already visible in KPI space in Table 6. This figure rests on a single family and one requirement vector, and is reported as an indication of magnitude rather than a characterized result. These measurements account for the modest Chamfer improvement of 8 to 13%, the weak KPI conditioning, the failure of every reconstruction route, and the observation that template deformation yields meshes visually indistinguishable from the template. The consequence for the roadmap is that solid mesh output is not blocked on a better reconstruction algorithm; it is blocked on a generation objective that produces normal-direction surface change.

### 7.4 Verified non-issues

Two alternative explanations were investigated and ruled out. Generated points lie on the manifold: the median distance to the nearest template surface point is 3.8 mm with a maximum of 29.7 mm, so no points occupy free space. And the raw per-index displacement statistic (median 5.4 mm, ninetieth percentile 57.8 mm) is a measurement artifact, since point i in two independently sampled clouds bears no correspondence; distance to surface is the correct measure.

## 8. Deployed Tool

The deployed pipeline accepts a requirement vector and a priority weighting, scores reachability across all 104 families, generates from a shortlist, evaluates each candidate with the surrogate, ranks by achieved rather than envelope-predicted performance, and returns the top candidates with three-dimensional views, KPI tables carrying resolvability labels, and downloadable point clouds. Figure 5 shows the requirement entry stage.

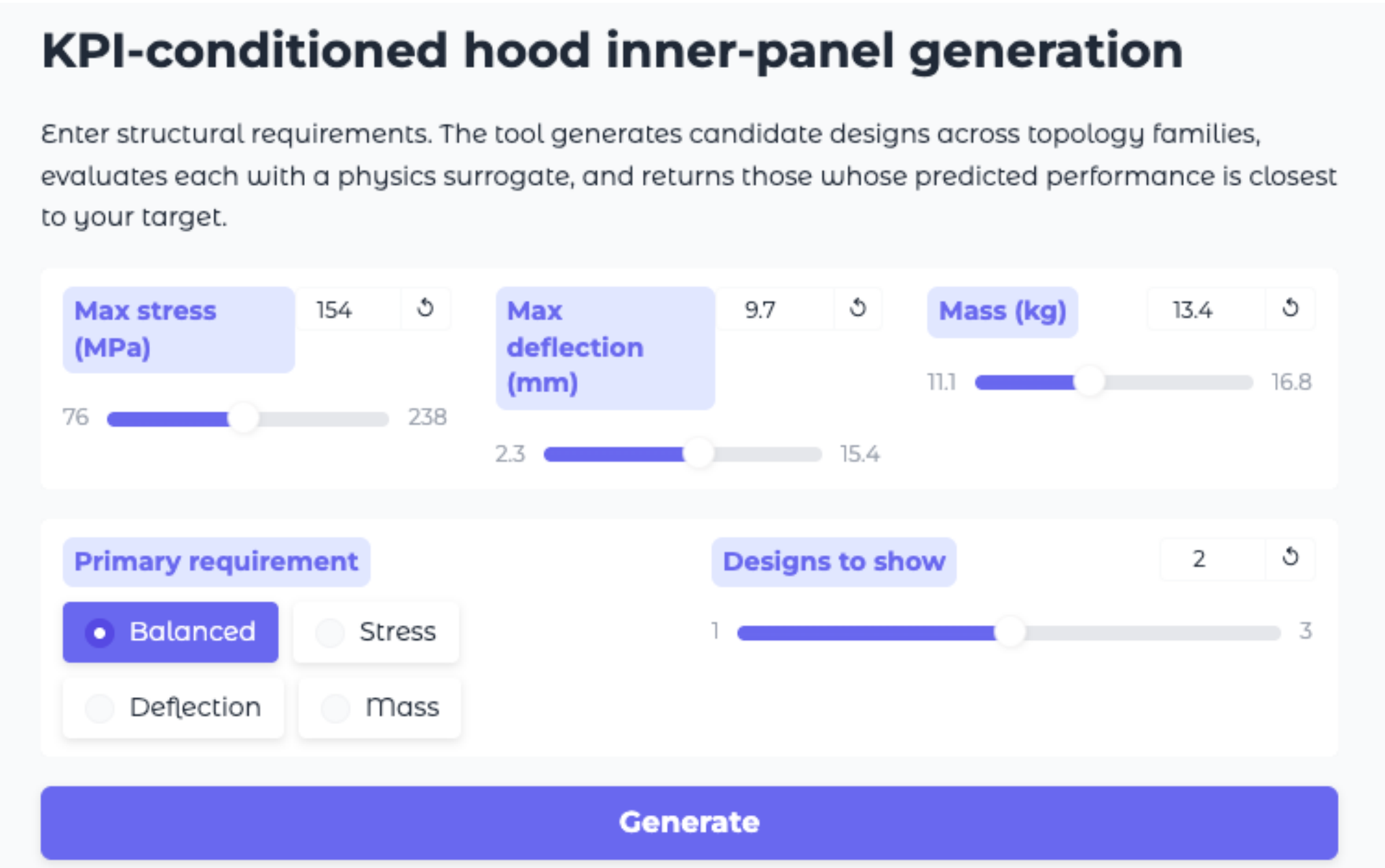

*Figure 5. Requirement entry in the deployed interface. Stress, deflection, and mass targets are set against sliders bounded by the range present in the training corpus. The primary-requirement selector supplies the weights $w^k$ of Section 5.1, and a separate control sets how many ranked candidates are returned.*

Ranking on achieved performance was a substantive correction rather than a refinement. In one test case a family whose envelope contained the target returned a 22.7% stress error; generating and scoring all shortlisted candidates demoted it automatically, and the highest-ranked result landed within 2.4%, inside surrogate error. Each KPI row reports the requested value, the surrogate estimate with its error band, the signed deviation marked against the surrogate floor, the family's measured range and mean from training data, and the resolvability verdict. The distinction between a meaningful estimate and a family-level one is carried per family and per KPI, so the tool never presents a number more precise than the evidence supports. Figure 6 shows this report for two families that differ in verdict while carrying comparable estimates, and Figure 7 the legend that accompanies it.

**2 best matches**

Generated 5 candidates across topology families; ranked by **achieved** performance with **balanced** weighted highest.

**skin_39 - 95 designs in training data**

| KPI | requested | predicted | delta | family range (mean) | resolution |
|---|---|---|---|---|---|
| Max stress (MPa) | 154.0 | **138.3** +/-10.2 | ! -10.2% | 105.9-190.4 (mean 140.3, +/-9.8%) | meaningful |
| Max deflection (mm) | 9.7 | **9.6** +/-0.5 | ok -0.8% | 7.7-10.7 (mean 9.6, +/-7.3%) | meaningful |
| Mass (kg) | 13.4 | **12.3** +/-0.2 | ! -8.5% | 11.5-12.4 (mean 12.0, +/-2.1%) | meaningful |

**skin_83 - 95 designs in training data**

| KPI | requested | predicted | delta | family range (mean) | resolution |
|---|---|---|---|---|---|
| Max stress (MPa) | 154.0 | **140.0** +/-10.3 | ! -9.1% | 143.0-170.7 (mean 152.3, +/-3.4%) | family-level only |
| Max deflection (mm) | 9.7 | **8.9** +/-0.5 | ! -8.2 % | 8.7-9.1 (mean 8.9, +/-0.8%) | family-level only |
| Mass (kg) | 13.4 | **13.9** +/-0.3 | ! +3.7 % | 13.7-14.4 (mean 14.0, +/-1.0%) | family-level only |

*Figure 6. Per-family KPI report for the two highest-ranked candidates returned against a representative requirement set. Each row carries the requested value, the surrogate estimate with its error band, the signed deviation, the family's measured range and mean from training data, and the resolvability verdict of Section 7.1. Family skin_39 resolves all three KPIs; skin_83 resolves none, and its estimates are therefore reported as family-level only despite comparable numerical precision.*

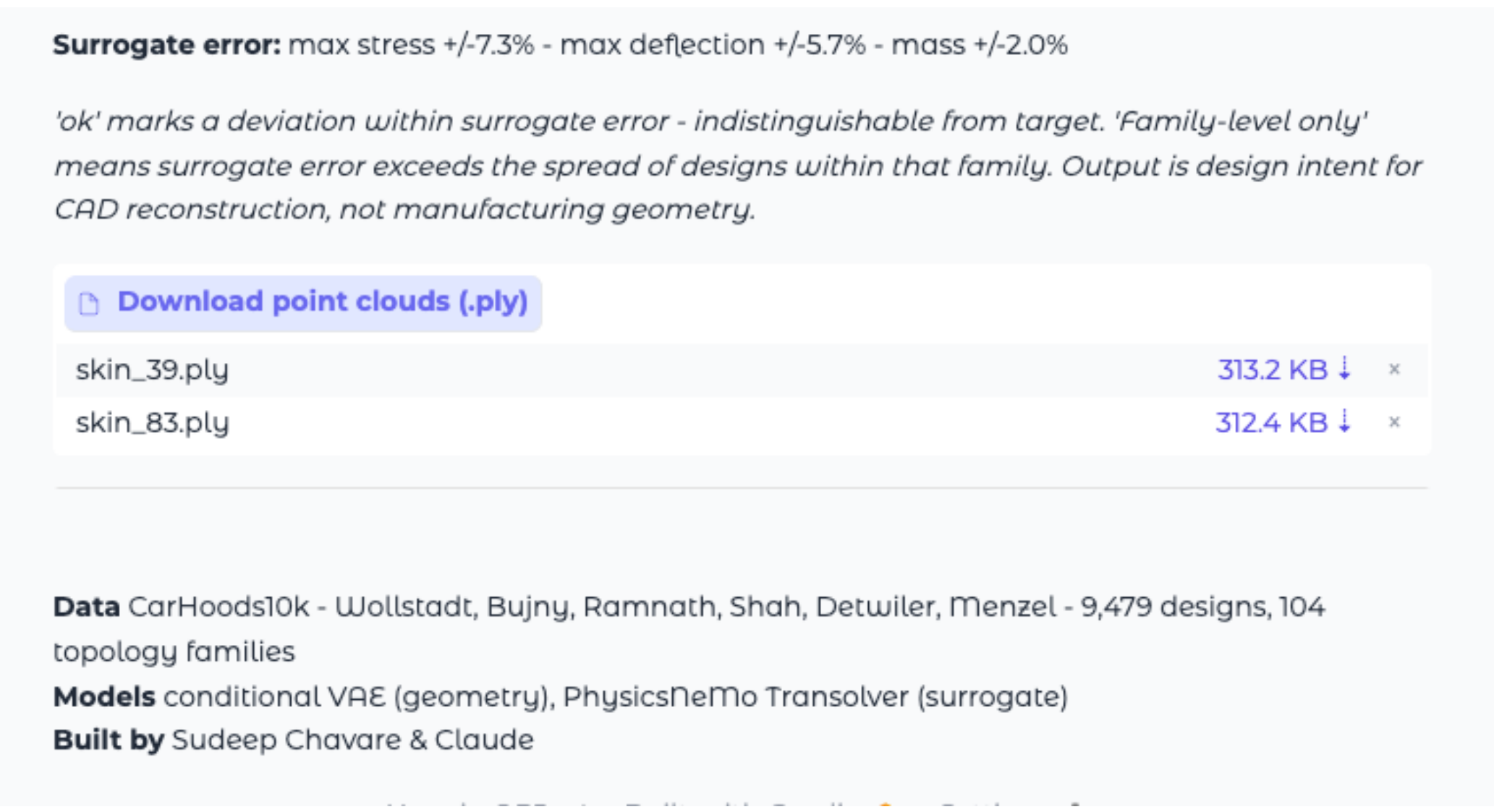

*Figure 7. Interpretation legend, download panel, and provenance footer accompanying every result. The surrogate error floor is stated explicitly, the 'ok' marker denotes a deviation indistinguishable from the target at that floor, and generated geometry is delivered as point clouds in PLY format.*

Output is delivered as point clouds colored by distance from the family reference. A returned candidate is shown in Figure 8. Solid mesh output is deferred per Section 7.3. This matches the framing adopted throughout: the deliverable is design intent for CAD reconstruction, not manufacturing geometry. The application runs as a Gradio interface on CPU inference, with a total artifact size of 13.1 MB comprising model weights, 104 reference point clouds, family envelope statistics, and configuration.

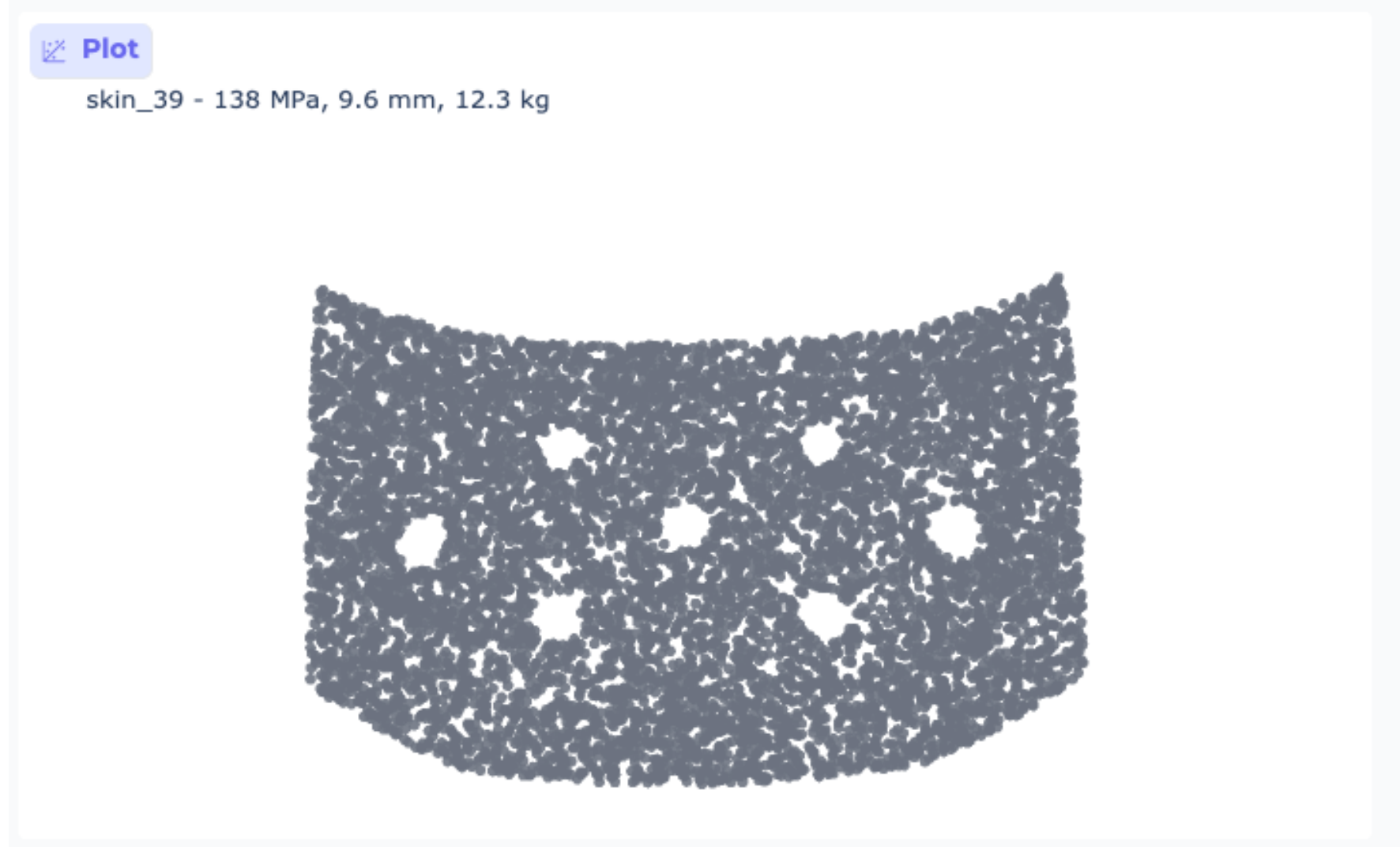

*Figure 8. Three-dimensional view of a returned candidate in the deployed interface, labeled with its surrogate-estimated KPIs.*

## 9. Limitations and Future Work

The most consequential limitation is the one identified in Section 7.3. Because the Chamfer objective permits tangential redistribution, the model produces little normal-direction surface change, which bounds both geometric fidelity and KPI conditioning strength. Three remedies are candidates: normal-direction supervision, in which the decoder predicts a scalar offset field along template normals and tangential freedom is eliminated by construction; a repulsion or spacing-regularization term penalizing redistribution; and an Earth Mover approximation, which enforces correspondence where Chamfer does not. This is the highest-value single change, since solid mesh output, stronger conditioning, and larger geometric change all depend on it. Once substantial normal deformation is achieved, template deformation becomes viable, as it already preserves inherited topology and crisp edges and failed only for want of real surface motion to transfer.

Reducing stress error below approximately 4% would make within-family estimates meaningful for most families rather than 45% of them. Options include field-level supervision, for which the per-token path has been retained; higher sampling density targeted specifically at fillets; and richer input features such as local curvature. Independently, high-fidelity finite element analysis on generated designs remains the only instrument with sufficient resolution to verify per-design KPI achievement; a single-family study of approximately 20 designs would settle definitively what the surrogate cannot. Separately, the error band the tool reports is a global test-set statistic rather than a per-design estimate; ensemble and Gaussian-process approaches to surrogate uncertainty quantification [14] would let the band vary with the design being evaluated and would sharpen the resolvability test of Section 7.1.

Two scope limits should be stated plainly. The pipeline cannot produce topologies outside the 104 families present in the data; latent interpolation between family embeddings is a natural extension, though validation would be difficult absent ground truth for topologies that do not exist. And the evaluation tests generalization to unseen variants of known families, not to unseen topologies, which follows from the deployment scenario but bounds the generality of the reported results.

## 10. Conclusions

A two-stage pipeline for KPI-conditioned generation of hood inner panel geometry was developed and deployed, comprising reachability-based topology retrieval, a conditional autoencoder operating on point clouds, and a Transolver surrogate for performance estimation. The surrogate reaches test errors of 7.34%, 5.70%, and 1.96% on stress, deflection, and mass. The generative model improves on the family reference design in 103 of 104 families, outperforms KPI-based retrieval, and attains 68% of the retrieval-oracle bound.

Three findings constrain the interpretation of these numbers, and are the more durable contribution. Surrogate error exceeds within-family KPI spread for most families, so per-design prediction is verifiable for only a minority; the tool reports this per family and per KPI rather than presenting uniform confidence. Surrogate-in-the-loop conditioning collapsed the model's response to its conditioning signal, for the diagnosable reason that supervision noisier than the signal it supervises drives a conditional model to its mean. And Chamfer-trained point-cloud generation was measured to redistribute points tangentially at 6.2 times the rate it deforms the surface normally, with zero net normal motion, a finding that explains the modest geometric improvement, the weak conditioning, and the failure of every mesh reconstruction route attempted.

The practical implication is that solid geometry output from this class of model is blocked on the training objective rather than on reconstruction technique. The methodological implication is that surrogate resolution should be characterized against the within-class variation a task requires before the surrogate is used either to validate generation or to supervise it.

## Acknowledgements

This work was conducted in the author's personal capacity, independent of any employer, using publicly available data and free-tier cloud compute. The author acknowledges Satchit Ramnath, Jami J. Shah, Patricia Wollstadt, Mariusz Bujny, Stefan Menzel, and Duane Detwiler of the Ohio State University and Honda Research Institute, responsible for producing and publicly releasing the CarHoods10k dataset, without which this work would not have been possible, and the NVIDIA PhysicsNeMo team for the open-source Transolver implementation. Anthropic's Claude was used to assist with pipeline design, diagnostic methodology, code development, and manuscript drafting; all results were generated, verified, and interpreted by the author.

## References

[1] Wu, H., Luo, H., Wang, H., Wang, J., and Long, M., "Transolver: A Fast Transformer Solver for PDEs on General Geometries," Proceedings of the 41st International Conference on Machine Learning (ICML), 2024. arXiv:2402.02366.

[2] Adams, C., Ranade, R., Cherukuri, R., and Choudhry, S., "GeoTransolver: Learning Physics on Irregular Domains Using Multi-scale Geometry Aware Physics Attention Transformer," arXiv preprint arXiv:2512.20399, 2025.

[3] Nabian, M. A., Chavare, S., Akhare, D., Ranade, R., Cherukuri, R., and Tadepalli, S., "Automotive Crash Dynamics Modeling Accelerated with Machine Learning," arXiv preprint arXiv:2510.15201, 2025. Also published as SAE Technical Paper 2026-01-0568, WCX SAE World Congress Experience, Detroit, MI, April 14, 2026. https://doi.org/10.4271/2026-01-0568.

[4] Akhare, D., Nabian, M. A., Adams, C., Chavare, S., and Choudhry, S., "High-Fidelity Industrial Crash Dynamics Prediction via Geometry-Aware Operator Learning with Memory-Efficient Low-Rank Attention," arXiv preprint arXiv:2605.27758, 2026.

[5] Ramnath, S., Shah, J. J., Wollstadt, P., Bujny, M., Menzel, S., and Detwiler, D., "OSU-Honda Automobile Hood Dataset (CarHoods10k)" [Dataset], Dryad, 2022. https://doi.org/10.5061/dryad.2fqz612pt.

[6] Wollstadt, P., Bujny, M., Ramnath, S., Shah, J. J., Detwiler, D., and Menzel, S., “CarHoods10k: An Industry-Grade Data Set for Representation Learning and Design Optimization in Engineering Applications,” IEEE Transactions on Evolutionary Computation, Special Issue on Benchmarking Sampling-Based Optimization Heuristics, 26(6):1221–1235, 2022. https://doi.org/10.1109/TEVC.2022.3147013.

[7] Sharma, V., Ganesh, H. J., Akram, M., Liu, W., and Raman, V., “AutoHood3D: A Multi-Modal Benchmark for Automotive Hood Design and Fluid–Structure Interaction,” NeurIPS 2025 Workshop on Machine Learning and the Physical Sciences (ML4PS), San Diego, CA, 2025. arXiv:2511.05596.

[8] Qi, C. R., Su, H., Mo, K., and Guibas, L. J., "PointNet: Deep Learning on Point Sets for 3D Classification and Segmentation," Proceedings of the IEEE Conference on Computer Vision and Pattern Recognition (CVPR), 2017.

[9] Kingma, D. P. and Welling, M., "Auto-Encoding Variational Bayes," International Conference on Learning Representations (ICLR), 2014.

[10] Sohn, K., Lee, H., and Yan, X., "Learning Structured Output Representation Using Deep Conditional Generative Models," Advances in Neural Information Processing Systems 28 (NeurIPS), 2015.

[11] Kazhdan, M., Bolitho, M., and Hoppe, H., "Poisson Surface Reconstruction," Eurographics Symposium on Geometry Processing, 2006.

[12] Bernardini, F., Mittleman, J., Rushmeier, H., Silva, C., and Taubin, G., "The Ball-Pivoting Algorithm for Surface Reconstruction," IEEE Transactions on Visualization and Computer Graphics 5(4):349-359, 1999.

[13] Ramnath, S., Haghighi, P., Kim, J. H., Detwiler, D., Berry, M., Shah, J. J., Aulig, N., Wollstadt, P., and Menzel, S., "Automatically Generating 60,000 CAD Variants for Big Data Applications," Proceedings of the ASME 2019 IDETC/CIE, Vol. 1: 39th Computers and Information in Engineering Conference, Anaheim, CA, August 18–21, 2019, Paper V001T02A006. https://doi.org/10.1115/DETC2019-97378.

[14] Chavare, S. and Mourelatos, Z., "Uncertainty Quantification in Machine Learning Using an Ensemble Approach with Gaussian Process Regression," WCX SAE World Congress Experience, Detroit, MI, April 8, 2025. https://doi.org/10.4271/2025-01-8199.

[15] NVIDIA Corporation, "PhysicsNeMo: An Open-Source Framework for Physics-Based Machine Learning," https://github.com/NVIDIA/physicsnemo, 2026.

## Appendix A. Training Failure Modes

Three failures preceded a working generative model. Each had a specific and transferable cause, and they are recorded here because the symptoms are easily misread as data or capacity limitations.

*A.1 Chamfer distance measuring sampling noise*

Initial runs subsampled 2,048 of 8,192 points independently from each cloud at every evaluation. A cloud compared against itself scored 0.0255, so the noise floor exceeded the real geometric signal of 0.008 by a factor of three. Training loss remained flat and the latent variable collapsed, because no gradient direction improved the metric. The correction was to compute Chamfer distance on full clouds; the self-comparison then scored 0.00007.

*A.2 Dead gradients from unnormalized ReLU stacks*

With ReLU activations and no normalization in the decoder, the final pre-activation went negative everywhere, so the output convolution received an identically zero input. Gradient norms were exactly zero for every layer except the output bias, and the model returned the template unchanged for 140 epochs. Group normalization with GELU activations resolved this. A related trap: initializing the output layer to exact zeros, intended to start the residual decoder at identity, blocks gradient flow to every upstream layer, since the upstream gradient is proportional to those zero weights. Small random initialization achieves near-identity behavior without severing the gradient path.

*A.3 Uninformative template pairing*

Templates were initially drawn at random from within the same family, so the same target appeared paired with many different starting geometries. The template therefore carried no information about the answer, only variance. Fixing one reference design per family resolved this and additionally aligned training with inference conditions, where a fixed reference is used.